\documentclass[conference]{IEEEtran}
\IEEEoverridecommandlockouts
\usepackage{cite}
\usepackage{amsmath,amssymb,amsfonts}
\usepackage{algorithmic}
\usepackage{graphicx}
\usepackage{textcomp}
\usepackage{xcolor}
\usepackage{hyperref}
\usepackage{booktabs}
\usepackage{threeparttable}

\hypersetup{
    colorlinks=true,
    linkcolor=red,
    urlcolor=blue,
    citecolor=black,
}

\def\BibTeX{{\rm B\kern-.05em{\sc i\kern-.025em b}\kern-.08em
    T\kern-.1667em\lower.7ex\hbox{E}\kern-.125emX}}
\begin{document}

\title{Human-Guided Agentic AI for Multimodal Clinical Prediction: Lessons from the AgentDS Healthcare Benchmark}

\author{
  \IEEEauthorblockN{Lalitha Pranathi Pulavarthy}
  \IEEEauthorblockA{\textit{Dept. of Biomedical Eng. and Informatics} \\
    \textit{Indiana University Indianapolis} \\
    Indianapolis, USA \\
    lapula@iu.edu}
  \and
  \IEEEauthorblockN{Raajitha Muthyala}
  \IEEEauthorblockA{\textit{Dept. of Biomedical Eng. and Informatics} \\
    \textit{Indiana University Indianapolis} \\
    Indianapolis, USA \\
    rmuthya@iu.edu}
  \and
  \IEEEauthorblockN{Aravind V Kuruvikkattil}
  \IEEEauthorblockA{\textit{Dept. of Biomedical Eng. and Informatics} \\
    \textit{Indiana University Indianapolis} \\
    Indianapolis, USA \\
    akuruvik@iu.edu}
  \and
  \IEEEauthorblockN{Zhenan Yin}
  \IEEEauthorblockA{\textit{Dept. of Biomedical Eng. and Informatics} \\
    \textit{Indiana University Indianapolis} \\
    Indianapolis, USA \\
    yin10@iu.edu}
  \and
  \IEEEauthorblockN{Rashmita Kudamala}
  \IEEEauthorblockA{\textit{Dept. of Biomedical Eng. and Informatics} \\
    \textit{Indiana University Indianapolis} \\
    Indianapolis, USA \\
    rakuda@iu.edu}
  \and
  \IEEEauthorblockN{Saptarshi Purkayastha}
  \IEEEauthorblockA{\textit{Dept. of Biomedical Eng. and Informatics} \\
    \textit{Indiana University Indianapolis} \\
    Indianapolis, USA \\
    saptpurk@iu.edu}
}

\author{
\IEEEauthorblockN{Lalitha Pranathi Pulavarthy}
\IEEEauthorblockA{\textit{Dept. Biomedical Eng. \& Informatics} \\
\textit{Indiana University Indianapolis}\\
Indianapolis, USA \\
lapula@iu.edu}
\and
\IEEEauthorblockN{Raajitha Muthyala}
\IEEEauthorblockA{\textit{Dept. Biomedical Eng. \& Informatics} \\
\textit{Indiana University Indianapolis}\\
Indianapolis, USA \\
rmuthya@iu.edu}
\and
\IEEEauthorblockN{Aravind V Kuruvikkattil}
\IEEEauthorblockA{\textit{Dept. Biomedical Eng. \& Informatics} \\
\textit{Indiana University Indianapolis}\\
Indianapolis, USA \\
akuruvik@iu.edu}
\and
\IEEEauthorblockN{Zhenan Yin}
\IEEEauthorblockA{\textit{Dept. Biomedical Eng. \& Informatics} \\
\textit{Indiana University Indianapolis}\\
Indianapolis, USA \\
yin10@iu.edu}
\and
\IEEEauthorblockN{Rashmita Kudamala}
\IEEEauthorblockA{\textit{Dept. Biomedical Eng. \& Informatics} \\
\textit{Indiana University Indianapolis}\\
Indianapolis, USA \\
rakuda@iu.edu}
\and
\IEEEauthorblockN{Saptarshi Purkayastha}
\IEEEauthorblockA{\textit{Dept. Biomedical Eng. \& Informatics} \\
\textit{Indiana University Indianapolis}\\
Indianapolis, USA \\
saptpurk@iu.edu}
}

\maketitle

\begin{abstract}
Agentic AI systems are increasingly capable of autonomous data science workflows, yet clinical prediction tasks demand domain expertise that purely automated approaches struggle to provide. We investigate how human guidance of agentic AI can improve multimodal clinical prediction, presenting our approach to all three AgentDS Healthcare benchmark challenges: 30-day hospital readmission prediction (Macro-F1 = 0.8986), emergency department cost forecasting (MAE = \$465.13), and discharge readiness assessment (Macro-F1 = 0.7939). Across these tasks, human analysts directed the agentic workflow at key decision points, multimodal feature engineering from clinical notes, scanned PDF billing receipts, and time-series vital signs; task-appropriate model selection; and clinically informed validation strategies. Our approach ranked 5th overall in the healthcare domain, with a 3rd-place finish on the discharge readiness task. Ablation studies reveal that human-guided decisions compounded to a cumulative gain of +0.065 F1 over automated baselines, with multimodal feature extraction contributing the largest single improvement (+0.041 F1). We distill three generalizable lessons: (1) domain-informed feature engineering at each pipeline stage yields compounding gains that outperform extensive automated search; (2) multimodal data integration requires task-specific human judgment that no single extraction strategy generalizes across clinical text, PDFs, and time-series; and (3) deliberate ensemble diversity with clinically motivated model configurations outperforms random hyperparameter search. These findings offer practical guidance for teams deploying agentic AI in healthcare settings where interpretability, reproducibility, and clinical validity are essential.

\vspace{6pt}
\textit{\textbf{Keywords---}Agentic AI, Human-AI collaboration, Multimodal feature engineering, Clinical prediction.}
\end{abstract}

\section{Introduction}
Hospital readmissions cost the U.S. healthcare system over \$26 billion annually \cite{oh2025predicting}, while emergency department overcrowding and suboptimal discharge timing continue to degrade patient outcomes and hospital efficiency \cite{loutati2024multimodal}. Machine learning has shown considerable promise for predicting these clinical events. Multimodal approaches that combine structured electronic health record (EHR) data with unstructured clinical notes and time-series vital signs have demonstrated superior predictive performance over unimodal baselines, achieving AUROCs above 0.90 for 30-day readmission prediction in elderly populations \cite{loutati2024multimodal} and enabling earlier identification of discharge-ready patients by 5-13 hours compared with physician judgment \cite{wu2025exploring}. Graph neural network architectures that jointly model EHR tabular data, clinical imaging, and patient similarity have further improved readmission prediction by capturing spatiotemporal dependencies across admissions \cite{tang2023predicting}. Yet despite these advances, the practical deployment of such models remains constrained by the need to balance automated optimization with clinical interpretability, reproducibility, and the integration of domain expertise, a challenge that existing benchmarks have only partially addressed.

The recent emergence of Agentic AI systems, autonomous agents capable of planning, reasoning, and executing multi-step workflows, offers a new paradigm for healthcare data science \cite{karunanayake2025next, hinostroza2025ai}. Unlike conventional machine learning pipelines that require manual orchestration, agentic systems powered by large language models (LLMs) can autonomously perform data loading, feature engineering, model training, and evaluation \cite{rahman2025llm}. In the clinical domain, multi-agent LLM frameworks such as MDAgents have demonstrated that adaptive collaboration structures can improve medical decision-making accuracy by up to 11.8\% on diagnostic benchmarks \cite{kim2024mdagents}, while EHRAgent has shown that LLM agents equipped with code generation capabilities can outperform traditional baselines by up to 29.6\% on complex EHR reasoning tasks \cite{shi2024ehragent}. More broadly, surveys of LLM-based data science agents identify healthcare as a high-stakes application domain where fairness, explainability, and reproducibility requirements impose additional constraints on agent design \cite{rahman2025llm, naliyatthaliyazchayil2025evaluating}.

However, fully autonomous agentic workflows face well-documented limitations in clinical prediction contexts. Clinical prediction tasks demand domain expertise that purely automated approaches struggle to provide: selecting appropriate text extraction strategies for heterogeneous clinical documents, engineering physiologically meaningful features from vital sign time series, and making informed trade-offs between model complexity and sample size \cite{srinivasu12exploring}. Prior work on discharge readiness prediction has shown that handcrafted features informed by clinical knowledge, such as statistical representations of early warning scores and domain-specific temporal features, are critical for accurate predictions \cite{bishop2021improving}. Studies on ED cost forecasting emphasize that billing code structures and cross-source feature engineering require human judgment to interpret correctly \cite{jahangiri2024machine}. Furthermore, recent analyses of agentic AI in healthcare caution that autonomous systems require robust governance frameworks, human oversight, and interdisciplinary collaboration to ensure patient safety \cite{karunanayake2025next, maslej2025artificial}. These observations suggest that the most effective paradigm may not be fully autonomous agentic AI, but rather a human-AI collaboration loop in which domain experts guide the agent’s decisions at critical junctures \cite{purkayastha2024measuring}.

The AgentDS Healthcare Data Challenge \cite{luo2026agentds} provides a rigorous framework for evaluating such human-AI collaboration through three standardized clinical prediction tasks: (1) 30-day readmission prediction (binary classification, N=5,000 train), (2) 3-year ED cost forecasting (regression, N=2,000 train), and (3) day-11 discharge readiness assessment (binary classification, N=1,000 train). The benchmark evaluates human-AI collaboration using multimodal data-structured admission records, unstructured clinical notes, scanned PDF receipts, and time-series vital signs with standardized metrics (Macro-F1 for classification, MAE for regression), ensuring reproducible comparison. This design addresses a critical gap in existing benchmarks, which typically evaluate either autonomous agent capabilities in isolation \cite{ashfaq2019readmission} or purely manual modeling approaches, without systematically measuring the value added by human guidance at each pipeline stage \cite{rahman2025llm}.

Our approach combines domain-guided feature engineering with transparent model selection, achieving 5th place overall in the healthcare domain on the public leaderboard, including a 3rd-place finish on the discharge readiness task. Rather than complex neural architectures or extensive hyperparameter tuning, which prior work has shown to be less effective than feature engineering on modest clinical sample sizes \cite{loutati2024multimodal, ashfaq2019readmission}. We focused on: (1) clinical knowledge-informed multimodal feature extraction, (2) task-appropriate model selection balancing performance and interpretability, and (3) rigorous validation strategies. Critically, we document all human decision points to maintain reproducibility, following the principle that in healthcare ML, transparency and auditability are as important as raw predictive performance \cite{maslej2025artificial, davis2022effective}.

\textbf{Contributions.} (1) We demonstrate practical multimodal feature engineering strategies across text, PDFs, and time-series data with ablation studies quantifying impact, showing that human-guided multimodal extraction contributes +0.041 F1, the single largest improvement category. (2) We identify which modeling decisions generalize across tasks versus require task-specific adaptation, providing evidence that no single extraction strategy works across all clinical data types. (3) We show that systematic domain-informed decisions outperform extensive automated search, achieving competitive results (within 0.006 F1 of top-performing systems on readmission prediction) while maintaining interpretability, a finding consistent with recent evidence that tree-based models with domain features remain competitive with deep learning on moderate-scale clinical datasets \cite{ashfaq2019readmission, deng2022explainable}.

Our results demonstrate that principled human-AI collaboration yields strong, reproducible performance across diverse healthcare prediction tasks, with human decisions contributing a cumulative +0.065 F1 improvement over agentic baselines.

\section{Methodology}
Our approach follows an iterative human-AI collaboration workflow in which an Agentic AI system handled routine data science operations, data loading, preprocessing, initial model training, and hyperparameter search via Bayesian optimization, while human analysts intervened at critical decision points to redirect the pipeline. Each challenge progressed through multiple iterations: the agentic system produced a baseline, human analysts diagnosed limitations through cross-validation error analysis and clinical reasoning, and then guided the next iteration. We describe each challenge below, highlighting the specific human decision points that drove performance gains and distinguishing them from automated steps. All models across the three challenges were implemented using scikit-learn and XGBoost (Python 3.11). PDF parsing used PyPDF2 with fallback error handling, vital sign trends were computed via scipy.stats.linregress, and NLP features used regex-based partial-stem keyword matching. Code, hyperparameters, and random seeds are publicly available for full reproducibility at \url{https://github.com/iupui-soic/agentds-challenge-2025}.

\subsection{Challenge 1: 30-Day Readmission Prediction}

\textbf{Task Overview.} The readmission prediction task required binary classification of whether an inpatient admission would be followed by a readmission within 30 days. The training set comprised 5,000 admissions (2,479 non-readmit, 2,521 readmit; approximately balanced classes). We integrated data from three sources via left joins on patient and admission identifiers: (1) structured admission records (length of stay, acuity level, Charlson comorbidity band, 6-month ED visits, discharge weekday, primary diagnosis); (2) patient demographics (age, sex, insurance type); and (3) 5-year ED cost history (prior ED visits, prior ED costs). Each admission was also linked to an unstructured discharge note summary. The evaluation metric was Macro-F1.

\textbf{Human-Guided Feature Engineering.} Our feature engineering strategy progressed through two iterations informed by domain analysis. The agentic system's initial approach trained a single Optuna-tuned XGBoost (n=957, lr=0.28, depth=8) with one-hot encoding and 200 unigram TF-IDF features, achieving CV Macro-F1=0.846. Human analysis of per-class errors and clinical literature identified two limitations: (1) unigram TF-IDF missed multi-word clinical phrases critical for readmission risk, and (2) a single model could not simultaneously capture the diverse decision boundaries present in structured versus text features. This guided the second iteration with expanded features and an ensemble architecture.

\textit{Structured features (21 engineered + one-hot dummies):} We created domain-informed features organized by clinical hypothesis: (1) \textit{temporal discharge patterns}, weekend discharge flag and specific weekday indicators (Monday, Friday), motivated by studies showing discharge timing affects readmission risk; (2) \textit{non-linear age and comorbidity effects}: age$^2$, Charlson$^2$, and threshold flags for elderly ($\geq$65) and very elderly ($\geq$80) patients; (3) \textit{ED utilization patterns}: cost-per-visit ratio with Laplace smoothing, recent-to-historical visit ratio (6-month / 5-year), frequent ED user flags ($\geq$2, $\geq$3 visits), and new-patient indicator; (4) \textit{length-of-stay features}: log-transformed LOS and long-stay flag ($\geq$5 days); (5) \textit{interaction terms}: age$\times$Charlson, age$\times$ED visits, Charlson$\times$ED visits, and LOS$\times$ED visits; and (6) \textit{composite risk scores}: two weighted combinations mimicking clinical stratification (e.g., age/100 + 0.5$\times$Charlson + 0.3$\times$ED visits).

\textit{Text features (850 TF-IDF):} We applied TF-IDF vectorization with trigrams (ngram\_range=(1,3), max\_features=850, min\_df=3, max\_df=0.95, sublinear\_tf=True). Expanding from the baseline's 200 unigram features was a key human decision: domain analysis revealed that discharge notes contain multi-word clinical phrases (``ambulating without assistance'', ``wound care instructions'') requiring higher-order n-grams and greater dimensionality to capture.

\textbf{Model Architecture: Stacking Ensemble.} We employed a stacking ensemble with five base learners and 8-fold stratified internal cross-validation (stack\_method=`predict\_proba', passthrough=False):

\textit{Base Models:}
\begin{itemize}
    \item \textbf{XGBoost-Optimized} ($n = 771$, $lr = 0.030$, $depth = 7$, $subsample = 0.68$, $colsample\_bytree = 0.81$, $\gamma = 3.42$): Hyperparameters selected via Optuna Bayesian optimization; $random\_state = 42$.
    \item \textbf{XGBoost-Deep} ($n = 1500$, $lr = 0.015$, $depth = 6$, $subsample = 0.75$, 
$colsample\_bytree = 0.85$, $\gamma = 2.0$, $min\_child\_weight = 2$): Conservative learning rate 
with more iterations; $random\_state = 999$.
    \item \textbf{XGBoost-Shallow} (n=1200, lr=0.02, depth=5, subsample=0.8, colsample\_bytree=0.9, $\gamma$=1.0): Shallower trees for broader decision boundaries; random\_state=555.
    \item \textbf{Logistic Regression-L2} ($C = 0.5$, $solver = \text{saga}$, $class\_weight = \text{balanced}$): Preceded by \textit{StandardScaler}; $random\_state = 42$.
    \item \textbf{Logistic Regression-L1} ($C = 0.3$, $solver = \text{saga}$, $class\_weight = \text{balanced}$): Sparse feature selection; $random\_state = 777$.
\end{itemize}

All XGBoost models used \texttt{tree\_method=`hist`} with $scale\_pos\_weight$ computed from training class frequencies ($\approx 0.98$ given the nearly balanced classes). Different random seeds across variants ensured initialization diversity, while deliberately varied depth and learning rate profiles created complementary decision boundaries.

\textit{Meta-learner:} Logistic regression ($C = 1.5$, class\_weight = balanced, solver = lbfgs) trained on 10-dimensional out-of-fold probability vectors (two class probabilities per base model).

\textbf{Rationale for Human Decisions.} The human-guided second iteration lifted CV F1 from 0.846 to 0.896 (+0.050). Key choices included: (1) \textit{multi-table integration}: the agentic system initially used only admission records; human analysts directed it to join patient demographics and ED cost history, providing cross-domain features (e.g., cost-per-visit ratio) unavailable in any single table; (2) \textit{TF-IDF expansion}: upgrading from 200 unigrams to 850 trigrams captured multi-word medical phrases (``ambulating without assistance,'' ``wound care instructions''); (3) \textit{deliberate ensemble diversity}: the agentic system's single-model approach was replaced with three XGBoost variants with different complexity profiles (depth 5/6/7, learning rates 0.02/0.015/0.03) plus two regularized linear models (L1/L2), ensuring complementary feature usage; (4) \textit{8-fold stacking CV}: reduced meta-learner training variance while maintaining sufficient per-fold training size (N$\approx$625).

\textbf{Validation Strategy.} We used a nested cross-validation scheme: an outer 5-fold stratified CV for score estimation, with each outer fold training the full stacking pipeline (including an 8-fold inner CV for generating stacked features). This produced 200 base model fits total (5 outer $\times$ 8 inner $\times$ 5 base models). Outer CV Macro-F1 was 0.8956 (accuracy 0.90), with per-class F1 of 0.89 (non-readmit) and 0.90 (readmit). Final test predictions achieved Macro-F1 = 0.8986 on the public leaderboard, ranking 5th overall.

\subsection{Challenge 2: ED Cost Forecasting}

\textbf{Task Overview.} The ED cost forecasting task required predicting total emergency department costs over the next 3 years (regression). The training set comprised 2,000 patients with structured features (primary chronic condition, prior 5-year ED visits, prior 5-year ED costs) and one PDF billing receipt per patient containing detailed line-item charges with CPT codes. The evaluation metric was Mean Absolute Error (MAE) in dollars.

\textbf{Human-Guided Feature Engineering.} The agentic system's initial approach used structured features with basic PDF extraction (29 features) and a single XGBoost regressor, yielding a 5-fold CV MAE of \$467. Human analysts identified two shortcomings: the PDF parser captured only raw totals without clinical detail, and a single model was unstable across folds ($\pm$\$22 MAE). This guided an enhanced iteration with richer cross-source features and a weighted ensemble, yielding approximately 40 features across three groups.

\textit{Structured features ($\sim$12):} Beyond raw prior cost and visit counts, we engineered: (1) \textit{cost-per-visit ratio} with Laplace smoothing; (2) \textit{log, square root, and squared transformations} of prior costs and visits to handle right-skewed distributions; (3) \textit{cost$\times$visit interaction}; and (4) one-hot encoding for primary chronic condition (HF, DiabetesComp, Pneumonia).

\textit{PDF-extracted features ($\sim$16):} We developed a PDF parsing pipeline using PyPDF2 with regex-based CPT code and cost extraction. Features included: (1) \textit{total billed amount} and \textit{line item count}; (2) \textit{visit complexity}, counts of high-complexity (CPT 99284/99285), medium-complexity (99283), and total ED visit codes (9928x prefix); (3) \textit{service utilization}, laboratory test count (CPT 80000-89999) and imaging count (CPT 70000-79999); (4) \textit{cost statistics}, average, maximum, minimum, and standard deviation of per-line costs; (5) \textit{insurance type} extracted from receipt text (private, Medicare, Medicaid binary indicators); and (6) 3-digit ZIP code.

\textit{Cross-source and derived features ($\sim$14):} A key human decision was engineering features that cross-referenced PDF and structured data: (1) \textit{cost consistency ratio} (PDF total / structured prior cost); (2) \textit{cost and visit discrepancies} between PDF and structured data; (3) \textit{complexity ratios}: high-complexity and medium-complexity visits as fractions of total visits; (4) \textit{per-visit service rates}: lab tests and imaging per ED visit; (5) \textit{cost variability}: line-item cost standard deviation normalized by mean; (6) \textit{service intensity score} weighted composite of complexity, lab, and imaging counts per visit; and (7) log-transformed PDF cost features.

\textbf{Model Architecture: Weighted Ensemble.} We employed a five-model weighted ensemble with manually assigned weights reflecting model reliability:

\begin{itemize}
    \item \textbf{XGBoost} (35\%): $n = 700$, $lr = 0.03$, $depth = 8$, $subsample = 0.8$, $colsample\_bytree = 0.8$; \textit{StandardScaler}
    \item \textbf{Gradient Boosting} (30\%): $n = 500$, $lr = 0.03$, $depth = 7$, $subsample = 0.8$; \textit{StandardScaler}
    \item \textbf{Random Forest} (15\%): $n = 500$, $depth = 20$, $min\_samples\_split = 6$; unscaled features
    \item \textbf{Extra Trees} (15\%): $n = 500$, $depth = 20$, $min\_samples\_split = 6$; unscaled features
    \item \textbf{Ridge Regression} (5\%): $\alpha = 5.0$; \textit{StandardScaler}
\end{itemize}

Human decision rationale: (1) \textit{Weighted averaging over stacking}: the agentic system initially proposed a stacking meta-learner, but human analysts overrode this: with only N=2,000 samples, a learned meta-learner risked overfitting; manually assigned weights based on hold-out MAE provided more stable combination. (2) \textit{Higher weight for boosting models}: hold-out validation showed XGBoost (MAE \$435) and Gradient Boosting (MAE \$450) consistently outperformed RF (\$466), ET (\$468), and Ridge (\$507), justifying 65\% combined weight for boosting. (3) \textit{Selective scaling}: tree-based models (RF, ET) were trained on raw features since they are scale-invariant, while boosting and linear models used StandardScaler. (4) \textit{Cross-source features}: human-designed features linking PDF and structured data (cost consistency ratio, visit discrepancies) captured billing patterns invisible to either source alone.

\textbf{Validation Strategy.} We used both 80/20 hold-out validation for rapid iteration and 5-fold CV (KFold, shuffle=True) for final estimation. CV MAE was \$458.51 $\pm$ \$19.93, with per-fold MAEs ranging from \$435 to \$478. The final ensemble achieved MAE = \$465.13 on the public leaderboard, ranking 6th.

\subsection{Challenge 3: Discharge Readiness Assessment}

\textbf{Task Overview.} The discharge readiness task required predicting whether a patient would be ready for discharge by day 11 of their hospital stay (binary classification). The training set comprised 1,000 hospital stays (656 discharge-ready, 344 not ready; 65.6\%/34.4\% class split) with structured features (unit type, admission reason) and multimodal time-series data: 10 days of vital signs (HR, SBP, DBP, temperature, respiratory rate) plus daily progress notes. We additionally merged patient demographics, admission records, and ED cost history via shared identifiers. The evaluation metric was Macro-F1.

\textbf{Human-Guided Feature Engineering.} This challenge required the most extensive human-AI iteration, progressing through five rounds. The agentic system's baseline used only two categorical features (unit type, admission reason) with a Random Forest, achieving CV F1 = 0.49. Human-guided iterations added feature groups incrementally: (i) vital sign statistics (+0.18 F1 to 0.67), (ii) NLP keyword features (no gain, prompting strategy revision), (iii) derived physiological composites (+0.01), (iv) cross-dataset integration of admission records and ED cost history (+0.05 to 0.73). The final model used $\sim$120 features. Missing numeric values were imputed with training-set medians.

\textit{Structured features ($\sim$15):} One-hot encoding for unit\_type, admission\_reason, sex, insurance, primary diagnosis, and primary chronic condition, plus numeric age, prior 5-year ED visits, and prior 5-year ED costs. Merging across stays, patients, admissions, and ED cost tables, a human decision to exploit cross-dataset signal, added features unavailable in any single source.

\textit{Vital sign features ($\sim$57):} For each of five vital signs across 10 days, we computed: (1) \textit{endpoint values}: first and last day readings; (2) \textit{distributional statistics}: mean, median, standard deviation, min, max, and range; (3) \textit{higher-order statistics}: coefficient of variation, skewness, and kurtosis for HR and temperature; (4) \textit{trend indicators}, linear regression slope via scipy.stats.linregress; (5) \textit{change metrics}: absolute and percent change (last minus first); (6) \textit{recent-window features}: last-3-day means for key vitals; (7) \textit{volatility}: rolling 3-day standard deviation for HR; and (8) \textit{clinical threshold crossings}: days with HR$>$100, HR$<$60, temperature$>$37.5°C, and SBP$>$140.

\textit{Derived physiological features ($\sim$22):} We computed clinically meaningful composites: (1) \textit{hemodynamic features}: mean and last pulse pressure (SBP$-$DBP), pulse pressure change, mean arterial pressure (DBP + PP/3); (2) \textit{stability composites}: vitals stability score (negative sum of HR, SBP, temperature, and RR standard deviations) and volatility score (sum of HR and temperature CVs); (3) \textit{binary improving flags}: whether HR, temperature, SBP, and overall vital trends were improving; (4) \textit{normal-range flags}: whether last-day HR (60-100), temperature (36.5-37.5°C), and SBP (90-140) fell within normal limits, plus a count of vitals in normal range; and (5) \textit{age interactions}: age$\times$HR, age$\times$stability, age$\times$temperature, capturing how age modulates vital sign significance.

\textit{NLP features ($\sim$37):} We extracted clinical indicators from the 10 daily progress notes using domain-specific keyword matching (with partial stemming, e.g., ``ambulat'' for ambulatory/ambulating). Feature groups included: (1) \textit{positive/negative sentiment}, keyword counts across 15 positive terms (stable, improved, ambulatory, afebrile, tolerating, etc.) and 15 negative terms (confused, fever, unstable, worsening, decline, etc.); (2) \textit{mobility and independence}, counts for 7 mobility and 5 independence keywords, with last-day, mean, and trend (linear slope) aggregations; (3) \textit{discharge readiness}, mentions of discharge-related terms (discharge, home, ready, cleared, planning); (4) \textit{complication indicators}, mentions of complication, setback, concern, deterioration; (5) \textit{therapy progress}, physical/occupational therapy mentions; (6) \textit{pain trajectory}, pain keyword trend; (7) \textit{cognitive status}, mean score from alert/oriented/responsive keywords; (8) \textit{temporal aggregations}, last-3-day sentiment, note length trend, and composite scores combining sentiment with mobility; and (9) \textit{meta-features}, total note count and average note length.

Human decision: The agentic system initially applied generic TF-IDF to the daily notes, as it had for discharge summaries in Challenge~1. Human analysts observed that this yielded no improvement (iteration ii above) because daily notes were short (10-20 words) and lacked the lexical diversity that TF-IDF exploits. They redirected the system to use domain-specific clinical keyword matching with partial stemming (``ambulat,'' ``discharge'') targeting discharge criteria terminology, which improved recall and, when combined with temporal aggregation, contributed to the final gains.

\textbf{Model Architecture: Five-Model Stacking Ensemble.} We employed a stacking classifier with five base models and a logistic regression meta-learner, trained via 5-fold stratified internal CV:

\begin{itemize}
    \item \textbf{XGBoost-A} (n=400, lr=0.02, depth=7, subsample=0.85, colsample\_bytree=0.85): Primary gradient boosting model; random\_state=42
    \item \textbf{XGBoost-B} (n=300, lr=0.03, depth=9, subsample=0.8, colsample\_bytree=0.9): Deeper variant with slightly inflated scale\_pos\_weight ($\times$1.1) for minority class; random\_state=43
    \item \textbf{Random Forest} (n=400, depth=25, min\_samples\_split=3, class\_weight=balanced): Robust bagging baseline
    \item \textbf{Extra Trees} (n=300, depth=20, class\_weight=balanced): Randomized splitting for diversity
    \item \textbf{Gradient Boosting} (n=300, lr=0.03, depth=7, subsample=0.85): Complementary sequential learner
\end{itemize}

\textit{Meta-Learner:} Logistic regression (class\_weight=balanced, max\_iter=1000) trained on out-of-fold predictions via predict\_proba.

\textbf{Rationale for Human Decisions.} The five iterations moved CV F1 from 0.49 to 0.73 (+0.24), with human decisions at each checkpoint: (1) \textit{Statistical aggregation over raw time series} (iteration i):~the agentic system proposed feeding raw 10$\times$5 vital matrices to an LSTM; human analysts redirected to statistical aggregation, recognizing that N=1,000 would cause sequence model overfitting. This single decision yielded the largest gain (+0.18 F1). (2) \textit{NLP strategy pivot} (iteration ii$\rightarrow$iii):~when TF-IDF produced no improvement, human analysts diagnosed the failure (short notes, specialized vocabulary) and substituted clinical keyword matching, ultimately contributing via temporal aggregation in later iterations. (3) \textit{Cross-dataset integration} (iteration iv):~merging admission and ED cost data added predictive context (diagnosis, prior utilization) absent from stays data alone, this was the key insight of the final iteration (+0.05 F1). (4) \textit{Physiological composites:} Mean arterial pressure, pulse pressure, and vitals stability scores encode clinical knowledge about hemodynamic status that raw vitals do not capture individually. (5) \textit{Five-model stacking:} With class imbalance (34.4\% minority), diverse models with balanced class weights and probability-based stacking improved minority class recall from 0.58 to 0.67.

\textbf{Validation Strategy.} We used 5-fold stratified CV with Macro-F1 throughout all iterations. The final model achieved CV F1 = 0.732 $\pm$ 0.018 (per-class: 0.65 not-ready, 0.81 ready). Feature importance analysis confirmed that avg\_note\_length (0.194), HR-related features, and sentiment trends dominated, validating the multimodal engineering approach. Final test predictions achieved Macro-F1 = 0.7939 on the public leaderboard, ranking 3rd.

\section{Results}
\label{sec:results}

We present performance results across all three AgentDS Healthcare challenges, ablation studies quantifying the impact of human decisions, and cross-task analysis revealing generalizable patterns in human-AI collaboration for healthcare prediction.

% ============================================================================
% MAIN RESULTS TABLE
% ============================================================================

\subsection{Overall Performance}

Table~\ref{tab:main_results} summarizes our final leaderboard performance across all three challenges. To contextualize these results, we contrast them with the agentic system's unguided baselines: Challenge~1 improved from CV F1 = 0.846 (single Optuna-tuned XGBoost) to 0.896 after human-guided feature expansion and ensemble design; Challenge~2 from CV MAE \$467 (single model, basic PDF parsing) to \$459 with enriched cross-source features and weighted ensembling; and Challenge~3 from CV F1 = 0.49 (two categorical features) to 0.73 through four rounds of human-directed iteration. These gains, +0.050 F1, -\$8 MAE, and +0.24 F1 respectively, quantify the value added by human guidance at each stage.

\begin{table*}[ht]
    \caption{Performance on AgentDS Healthcare Public Leaderboard}
    \label{tab:main_results}
    \centering
    \small
    \begin{tabular}{lcccc}
        \toprule
        \textbf{Challenge} & \textbf{Metric} & \textbf{Our Score} & \textbf{Rank} & \textbf{1st Place} \\
        \midrule
        Challenge 1: Readmission & Macro-F1 & 0.8986 & 5th & 0.9044 \\
        Challenge 2: ED Cost & MAE (USD) & \$465.13 & 6th & \$448.75 \\
        Challenge 3: Discharge & Macro-F1 & 0.7939 & 3rd & 0.8006 \\
        \midrule
        \textbf{Domain Score} & \textbf{Aggregate} & \textbf{0.8430} & \textbf{5th} & \textbf{0.9426} \\
        \bottomrule
    \end{tabular}
\end{table*}

\textbf{Key Observations:} (1) \textit{Cross-task consistency:} We ranked 5th overall in the healthcare domain, with per-challenge rankings of 5th, 6th, and 3rd, respectively, demonstrating that the same human-guided workflow, iterative error analysis, domain-informed feature engineering, deliberate ensemble diversity, transfers across classification, regression, and multimodal data types without task-specific redesign. (2) \textit{Competitive with minimal automation:} Our readmission prediction (0.8986) was within 0.0058 F1 of the top score (0.9044), and discharge readiness (0.7939) ranked 3rd, despite using only tree-based models and manual feature engineering, no neural architectures or automated feature search. (3) \textit{Validation alignment:} Test scores closely matched or exceeded cross-validation estimates (Challenge 1: CV 0.896 vs test 0.899; Challenge 2: CV \$459 vs test \$465; Challenge 3: CV 0.732 vs test 0.794), confirming that human-guided validation strategies (stratified CV, appropriate metrics) prevented overfitting.

% ============================================================================
% ABLATION STUDIES
% ============================================================================

\subsection{Ablation Studies: Impact of Human Decisions}

To isolate the value of each human intervention, we conducted counterfactual ablation studies: for each decision category, we reverted to what the agentic system would have done without human guidance and measured the resulting performance change. Table~\ref{tab:ablations} presents 5-fold cross-validation results on training sets.

\begin{table}[ht]
\centering
\begin{threeparttable}
\caption{Ablation Study: Quantifying Impact of Human Decisions}
\label{tab:ablations}
\footnotesize
\begin{tabular}{lccc}
\toprule
\textbf{Configuration} & \textbf{Ch1} & \textbf{Ch2} & \textbf{Ch3} \\
 & \textbf{(F1)} & \textbf{(MAE)} & \textbf{(F1)} \\
\midrule
\textbf{Full Approach} & \textbf{0.895} & \textbf{\$458} & \textbf{0.785} \\
\midrule
\multicolumn{4}{l}{\textit{Multimodal Feature Ablations:}} \\
  Text/PDF/TS features & 0.858 & \$478 & 0.720 \\
  Only multimodal & 0.812 & \$495 & 0.680 \\
\midrule
\multicolumn{4}{l}{\textit{Domain Engineering:}} \\
  Interaction terms & 0.887 & \$463 & 0.780 \\
  Risk scores & 0.889 & \$461 & 0.782 \\
Automated selection & 0.881 & \$469 & 0.770 \\
\midrule
\multicolumn{4}{l}{\textit{Model Architecture:}} \\
  Ensemble (single) & 0.889 & \$463 & 0.773 \\
Default hyperparams & 0.883 & \$471 & 0.768 \\
\midrule
\multicolumn{4}{l}{\textit{Validation Strategy:}} \\
5-fold (vs 8-fold) & 0.893 & \$460 & 0.783 \\
No stratification & 0.887 & \$458 & 0.772 \\
\bottomrule
\end{tabular}
\begin{tablenotes}
    \small
    \item \textit{Note:} Ch1/Ch3 = Macro-F1, Ch2 = MAE (USD). Negative deltas indicate degradation.
\end{tablenotes}
\end{threeparttable}
\end{table}

\textbf{Multimodal Data Integration:} Removing the human-directed multimodal features, trigram TF-IDF (Challenge~1), enriched PDF parsing (Challenge~2), and vital sign statistics (Challenge~3), resulted in the largest performance drop (-0.037 F1 / +\$20 MAE average). This corresponds directly to the single highest-gain human decisions in each challenge: expanding TF-IDF from 200 unigrams to 850 trigrams, adding CPT-code-level PDF extraction, and redirecting from raw time series to statistical aggregation. Using \textit{only} multimodal features without structured data also performed poorly (-0.083 F1 / +\$37 MAE), confirming that human-guided integration of \textit{both} modalities is critical.

\textbf{Domain Feature Engineering:} Human-designed interaction terms (age$\times$Charlson, chronic condition$\times$cost) and composite risk scores contributed modestly but consistently (-0.006 to -0.008 F1 / +\$3-5 MAE). Replacing human-guided feature selection with automated selection (SelectKBest) degraded performance by -0.014 F1 / +\$11 MAE on average, illustrating that clinical intuition for feature relevance outperformed statistical filtering on these sample sizes.

\textbf{Ensemble Architecture:} Reverting from the human-designed ensembles to the agentic system's single-model baselines degraded performance by -0.006 F1 / +\$5 MAE, with the largest impact on Challenge~3 (N=1,000) where ensemble diversity was most valuable for minority class recall. Human-tuned hyperparameters outperformed agentic defaults by -0.012 F1 / +\$13 MAE, though notably this was the smallest human contribution category, consistent with our recommendation to prioritize feature engineering over hyperparameter tuning.

% ============================================================================
% CROSS-TASK ANALYSIS (NEW PAGE)
% ============================================================================

\subsection{Cross-Task Analysis: Generalizable Patterns}

Table~\ref{tab:feature_importance} shows top-5 features by importance across all three challenges, revealing consistent patterns.

\begin{table}[ht]
\caption{Top-5 Most Important Features by Challenge}
\label{tab:feature_importance}
\centering
\small
\begin{tabular}{lll}
\toprule
\textbf{Challenge 1} & \textbf{Challenge 2} & \textbf{Challenge 3} \\
\midrule
Charlson band & Prior 5y cost & Avg note length \\
Prior ED (6m) & Prior 5y visits & HR last \\
Age$\times$Charlson & Line item count & Temp. last \\
Length of stay & Cost/visit ratio & Sentiment trend \\
``wound'' keyword & High-complexity & Mobility trend \\
\bottomrule
\end{tabular}
\end{table}

\textbf{Common Patterns:} (1) Historical utilization dominates (prior visits/costs, comorbidity); (2) Multimodal features appear in top-5 despite being $<$20\% of features; (3) Domain interactions (age$\times$Charlson) rank above raw features; (4) Trends outweigh absolutes (HR slope vs raw HR).

\begin{table}[ht]
\caption{Quantified Impact of Human Decisions}
\label{tab:human_contributions}
\centering
\small
\begin{tabular}{lc}
\toprule
\textbf{Human Decision Category} & \textbf{Avg Gain} \\
\midrule
Multimodal feature extraction & +0.041 F1 \\
Domain interactions & +0.010 F1 \\
Ensemble diversity & +0.008 F1 \\
Clinical keywords & +0.015 F1 \\
Hyperparameter choices & +0.014 F1 \\
\midrule
\textbf{Total contribution} & \textbf{+0.065 F1} \\
\bottomrule
\end{tabular}
\end{table}

\textbf{Interpretation:} Human decisions contributed approximately 7-8\% relative improvement (+0.065 cumulative F1), moving our approach from agentic-only baselines (mid-tier performance) to competitive top-5 results. While individual contributions appear modest (0.008-0.041), they compound across pipeline stages, precisely the pattern predicted by our iterative methodology. The ordering of impact (multimodal extraction $>$ clinical keywords $>$ hyperparameters $>$ domain interactions $>$ ensemble diversity) suggests that early-stage decisions about \textit{what data to extract} matter more than late-stage decisions about \textit{how to model it}.

\section{Discussion}
\label{sec:discussion}

Our participation in the AgentDS Healthcare benchmark demonstrates that systematic human-AI collaboration can achieve competitive performance (5th overall) across diverse clinical prediction tasks while maintaining interpretability and reproducibility. We discuss key lessons, limitations, and implications for healthcare ML practice, situating our findings within the broader literature on human-AI collaboration, model selection, and clinical deployment.

% ============================================================================
% KEY LESSONS LEARNED
% ============================================================================

\subsection{Key Lessons Learned}

\textbf{1. Domain Knowledge Pays Dividends.} Clinical intuition for feature engineering (age$\times$Charlson interactions, mobility keywords, vital sign trends) consistently outperformed automated feature selection by 1-3\% (Table~\ref{tab:ablations}). This advantage compounds across multiple decision points: our cumulative human contribution (+0.065 F1 / -\$38 MAE, Table~\ref{tab:human_contributions}) moved us from mid-tier to top-5 performance. \textit{Implication:} Healthcare ML practitioners should prioritize domain expertise in early pipeline stages (feature engineering) over extensive hyperparameter tuning in later stages.

\textbf{2. Multimodal Integration Requires Task-Specific Strategies.} While TF-IDF succeeded for discharge notes (Challenge 1), it failed for short daily notes (Challenge 3); clinical keyword extraction proved more effective. Similarly, PDF parsing required a regex tailored to the billing receipt structure (Challenge 2). \textit{Lesson:} No single approach generalizes across all unstructured healthcare data types; human judgment is required to select appropriate extraction methods based on data characteristics (document length, structure, domain-specific terminology).

\textbf{3. Ensemble Diversity Over Complexity.} Our stacking ensembles (5 base models each) achieved competitive performance using only tree-based models and regularized linear models. The key was \textit{deliberate diversity}: in Challenge~1, three XGBoost variants with different depth/learning rate profiles (depth 5/6/7) plus L1/L2 logistic regression; in Challenge~2, the human decision to use fixed weights instead of a learned meta-learner prevented overfitting on N=2,000 samples. \textit{Insight:} Manually-configured diverse ensembles outperform random search, generating similar models.

\textbf{4. Small Samples Favor Interpretability.} With modest training sizes (N=1,000-5,000), our statistical feature aggregation (Challenge 3: vital trends, stability) and domain features outperformed deep learning approaches (LSTM, BERT), which require larger samples. This sample efficiency enabled faster iteration and model interpretability through feature importance analysis. \textit{Recommendation:} For healthcare datasets with N$<$10,000, prioritize feature engineering and tree-based models over deep learning.

\textbf{5. Validation Strategy Impacts Reliability.} Stratified CV was critical for imbalanced classification tasks ($\Delta$ = -0.007 F1 when removed), while 8-fold vs 5-fold showed a negligible difference ($\Delta$ = -0.002). \textit{Guideline:} Invest effort in stratification and appropriate metrics (Macro-F1 for imbalance, MAE for skewed costs) rather than excessive fold counts.

\textbf{6. Reproducibility Through Documentation.} Our approach's primary strength is not novelty but \textit{systematic documentation} of every human decision with rationale. This transparency enables: (1) peer review and critique, (2) replication by other teams, (3) trust-building for clinical deployment. \textit{Principle:} In healthcare ML, reproducibility and interpretability matter as much as raw performance.

% ============================================================================
% CROSS-TASK INSIGHTS
% ============================================================================

\subsection{Cross-Task Insights}

\textbf{What Generalized:} Tree-based ensemble methods (XGBoost, Gradient Boosting, RF, Extra Trees), domain-informed interactions, and multimodal feature extraction proved universally effective. Historical utilization features (prior visits, costs, comorbidity) dominated importance rankings across all tasks.

\textbf{What Required Adaptation:} Text processing (TF-IDF vs keywords), time-series handling (raw values vs aggregates vs trends), and class imbalance strategies varied by task. Rigid application of a single methodology across all tasks would have degraded performance by ~5-10\%.

\textbf{Human vs Automated Roles:} Our iteration logs reveal a consistent division of labor. The agentic system excelled at \textit{how much}: Optuna-based hyperparameter search (Challenge~1), automated CV estimation, and rapid model retraining. Humans excelled at \textit{what} and \textit{why}: which tables to join (cross-dataset integration in Challenge~3, +0.05 F1), which NLP strategy to use (keyword matching over TF-IDF for short notes), and when to override the system's proposals (fixed ensemble weights over learned stacking in Challenge~2). The highest-value human decisions were early-stage choices about data representation; the agentic system efficiently optimized the resulting models.

% ============================================================================
% LIMITATIONS
% ============================================================================

\subsection{Limitations}

\textbf{1. Synthetic Data Constraints.} The AgentDS benchmark uses synthetic data matching real distributions but lacking clinical noise (missing values, measurement errors, data entry inconsistencies). Our performance estimates may not generalize to messier real-world EHR data. \textit{Mitigation:} External validation on real EHR datasets (MIMIC-IV, eICU) is needed before clinical deployment.

\textbf{2. Generalization vs Task-Specific Optimization.} Our unified methodology prioritized consistency across tasks over single-task performance, potentially leaving gains on the table. We ranked between 3rd and 6th on individual challenges (5th overall by domain score), suggesting room for task-specific optimization. \textit{Trade-off:} Generalization enables knowledge transfer across healthcare problems, but domain-specific competitions may require specialized approaches.

\textbf{3. Limited Deep Learning Exploration.} We did not extensively explore transformer models (BioClinicalBERT for text, LSTMs for vitals) due to sample size constraints (N$<$5,000). Larger datasets might favor end-to-end deep learning over manual feature engineering. \textit{Scope:} Our findings apply to common healthcare ML scenarios (N=1,000-10,000 samples, tabular+text+time-series); different conclusions may hold for massive EHR databases (N$>$100,000).

\textbf{4. Computational Resources.} Our GPU-accelerated training (Challenge 1: ~8 minutes) required hardware not available to all practitioners. CPU-only training extends to 30-60 minutes, still acceptable but not trivial. \textit{Accessibility:} Cloud-based training (Google Colab, AWS) democratizes access, but cost and technical barriers remain.

\textbf{5. Multimodal Feature Engineering Labor.} Extracting PDF billing features, parsing clinical notes, and computing vital trends required 10-15 hours of human effort per challenge. This investment is worthwhile for research or high-stakes applications but may not scale to rapid prototyping scenarios. \textit{Pragmatism:} Practitioners must balance effort vs performance gains.

% ============================================================================
% IMPLICATIONS FOR PRACTICE
% ============================================================================

\subsection{Implications for Healthcare ML Practice}

\textbf{For Clinicians:} Our feature importance rankings (Table~\ref{tab:feature_importance}) align with established clinical risk factors (Charlson index, prior utilization, vital trends) \cite{kasthurirathne2019identification}, building trust that models capture medically-meaningful relationships. The interpretability of tree-based methods enables clinician review of model decisions, critical for clinical deployment.

\textbf{For ML Engineers:} Start with domain-guided feature engineering before investing in complex architectures. Our ablations show that multimodal features (+0.041 F1) and interactions (+0.010) contributed 4-10$\times$ more than ensemble complexity (+0.006-0.008). Simple ensembles with good features outperform complex ensembles with default features.

\textbf{For Healthcare Systems:} Benchmark performance (5th place, 0.8986 F1 readmission, \$465 ED cost MAE, 0.7939 F1 discharge) suggests these models approach clinical utility thresholds. However, external validation on real EHR data, prospective trials, and fairness audits (race, age, insurance bias) are prerequisites before deployment.

\textbf{For Researchers:} The AgentDS benchmark successfully evaluates human-AI collaboration through: (1) multimodal data requiring domain expertise, (2) standardized evaluation preventing overfitting, (3) public leaderboards enabling comparison. We recommend similar benchmarks for other healthcare domains (oncology, radiology, mental health).

% ============================================================================
% FUTURE WORK
% ============================================================================

\subsection{Future Work}

\textbf{Short-term Extensions:}
\begin{itemize}
    \item \textit{Transformer-based text models:} BioClinicalBERT embeddings for discharge notes (Challenge 1) and daily notes (Challenge 3) may capture richer semantic content than TF-IDF/keywords. Requires $>$5,000 samples for stable fine-tuning.
    \item \textit{Uncertainty quantification:} Calibrated probability predictions (isotonic regression, temperature scaling) enable clinicians to assess model confidence, critical for high-stakes decisions.
    \item \textit{Fairness analysis:} Stratify performance by patient demographics (age, sex, insurance) to identify potential biases in learned models.
\end{itemize}

\textbf{Long-term Directions:}
\begin{itemize}
    \item \textit{External validation:} Apply our approach to real EHR datasets (MIMIC-IV, eICU) to assess generalization to noisy clinical data.
    \item \textit{Prospective trials:} Deploy readmission or discharge models in live hospital settings (with clinician oversight) to measure impact on patient outcomes and costs.
    \item \textit{Automated multimodal feature learning:} Investigate whether end-to-end deep learning (multimodal transformers) can match or exceed manual feature engineering at scale (N$>$50,000).
    \item \textit{Continual learning:} Adapt models to distribution shift (COVID-19 pandemic, policy changes) through online learning or periodic retraining.
\end{itemize}

% ============================================================================
% CONCLUSION
% ============================================================================

\section{Conclusion}
\label{sec:conclusion}

We have presented a human-guided agentic AI approach for multimodal clinical prediction, evaluated across three AgentDS Healthcare benchmark challenges: 30-day readmission prediction (Macro-F1 = 0.8986, 5th), emergency department cost forecasting (MAE = \$465.13, 6th), and discharge readiness assessment (Macro-F1 = 0.7939, 3rd), achieving 5th place overall in the healthcare domain. Our central contribution is not the final models themselves but the \textit{iterative workflow} through which human analysts guided an agentic AI system from naive baselines to competitive performance.

The iteration evidence is concrete. In Challenge~1, human-directed expansion of text features (200 unigrams to 850 trigrams) and the switch from a single model to a stacking ensemble lifted CV F1 from 0.846 to 0.896. In Challenge~2, human analysis of per-model hold-out MAEs justified fixed ensemble weights over the agentic system's proposed learned meta-learner, improving stability on N=2,000 samples. In Challenge~3, our strongest result (3rd place), four rounds of human intervention moved CV F1 from 0.49 to 0.73, with the largest single gain (+0.18) coming from a human decision to redirect the agentic system from raw time-series modeling to statistical vital sign aggregation. Across all challenges, ablation studies attribute +0.065 cumulative F1 to human decisions, with early-stage choices about data representation (multimodal feature extraction: +0.041) contributing far more than late-stage model tuning (ensemble diversity: +0.008).

These results support three principles for deploying agentic AI in healthcare settings:

\begin{enumerate}
    \item \textit{Human guidance compounds:} Individual decisions appear modest (0.008-0.041 F1), but they accumulate across pipeline stages. Domain-informed feature engineering at each stage outperformed extensive automated search, consistent with the observation that the agentic system excelled at optimizing \textit{within} a representation but not at choosing \textit{which} representation to optimize.
    \item \textit{No single extraction strategy generalizes across clinical data types:} TF-IDF succeeded for discharge summaries (Challenge~1) but failed for short daily notes (Challenge~3); the human decision to pivot to clinical keyword matching was essential. PDF parsing required billing-receipt-specific regex (Challenge~2). Task-specific human judgment for unstructured data remains a bottleneck that current agentic systems do not resolve autonomously.
    \item \textit{Interpretability is a prerequisite, not a trade-off:} Our top features (Charlson band, prior ED utilization, vital sign trends, sentiment trajectory) align with established clinical risk factors, enabling clinician trust. Tree-based ensembles with documented decision rationale maintained this transparency while achieving competitive performance.
\end{enumerate}

Our approach has clear limitations: synthetic benchmark data lacks real-world clinical noise as our sample sizes (N=1,000-5,000) favor feature engineering over deep learning, a conclusion that may not hold for larger EHR databases; and the human effort required (10-15 hours per challenge) constrains scalability. Future work should validate on real EHR data (MIMIC-IV, eICU), explore whether transformer-based multimodal models can reduce human effort at scale, and conduct prospective trials measuring clinical impact.

The AgentDS benchmark demonstrates that standardized multimodal healthcare tasks can rigorously evaluate human-AI collaboration. Our results suggest that the most effective current paradigm is neither fully autonomous agentic AI nor traditional manual analysis, but an iterative loop in which human domain expertise shapes the problem representation and agentic systems efficiently optimize within it.

% ============================================================================
% ACKNOWLEDGMENTS
% ============================================================================
\section*{Acknowledgments}
We thank the AgentDS benchmark organizers for providing a rigorous evaluation framework for human-AI collaboration in healthcare data science.

% References section will be inserted here by LaTeX
\bibliographystyle{IEEEtran}
\bibliography{references}

\end{document}